# SKINCURE: AN INNOVATIVE SMART PHONE-BASED APPLICATION TO ASSIST IN MELANOMA EARLY DETECTION AND PREVENTION

Omar Abuzaghleh [1], Miad Faezipour [2] and Buket D. Barkana [3]

[1]Department of Computer Science and Engineering, University of Bridgeport, Bridgeport, CT, USA
oabuzagh@bridgeport.edu

[2]Department of Computer Science and Engineering, Department of Biomedical Engineering, University of Bridgeport, Bridgeport, CT, USA
mfaezipo@bridgeport.edu

[3]Department of Electrical Engineering, University of Bridgeport, Bridgeport, CT, USA
bbarkana@bridgeport.edu

## ABSTRACT

*Melanoma spreads through metastasis, and therefore it has been proven to be very fatal. Statistical evidence has revealed that the majority of deaths resulting from skin cancer are as a result of melanoma. Further investigations have shown that the survival rates in patients depend on the stage of the infection; early detection and intervention of melanoma implicates higher chances of cure. Clinical diagnosis and prognosis of melanoma is challenging since the processes are prone to misdiagnosis and inaccuracies due to doctors' subjectivity. This paper proposes an innovative and fully functional smart-phone based application to assist in melanoma early detection and prevention. The application has two major components; the first component is a real-time alert to help users prevent skin burn caused by sunlight; a novel equation to compute the time for skin to burn is thereby introduced. The second component is an automated image analysis module which contains image acquisition, hair detection and exclusion, lesion segmentation, feature extraction, and classification. The proposed system exploits PH2 Dermoscopy image database from Pedro Hispano Hospital for development and testing purposes. The image database contains a total of 200 dermoscopy images of lesions, including normal, atypical, and melanoma cases. The experimental results show that the proposed system is efficient, achieving classification of the normal, atypical and melanoma images with accuracy of 96.3%, 95.7% and 97.5%, respectively.*

## 1. INTRODUCTION

Nowadays, skin cancer has been increasingly identified as one of the major causes of deaths. Research has shown that there are numerous types of skin cancers. Recent studies have shown that there are approximately three commonly known types of skin cancers. These include melanoma, basal cell carcinoma (BCC), and squamous cell carcinomas (SCC) [1]. However, melanoma has been considered as one of the most hazardous types in the sense that it is deadly, and its prevalence has enormously increased with time. Melanoma is a condition or a disorder that affects the melanocyte cells thereby impeding the synthesis of melanin [2]. A skin that has inadequate melanin is exposed to the risk of sunburns as well as harmful ultra-violet rays from the sun [3]. Researchers claim that the disease requires early intervention in order to be able to identify exact symptoms that will make it easy for the clinicians and dermatologists to prevent further infection. This disorder has been proven to be unpredictable. It is characterized by development of lesions in the skin that vary in shape, size, color and texture.

Though most people diagnosed with skin cancer have higher chances to be cured, melanoma survival rates are lower than that of non-melanoma skin cancer [4]. As more new cases of skin cancer are being diagnosed in the U.S. each year, an automated system to aid in the prevention and early detection is highly in-demand [5]. Following are the estimations of the American Cancer Society for melanoma in the United States for the year 2014 [6]:

• Approximately 76,100 new melanomas are to be diagnosed (about 43,890 in men and 32,210 in women).
• Approximately 9,710 fatalities are expected as a result of melanoma (about 6,470 men and 3,240 women).

For 30 years, more or less, melanoma rates have been increasing steadily. It is 20 times more common for white people to have melanoma than in African-Americans. Overall, during the lifetime, the risk of developing melanoma is approximately 2% (1 in 50) for whites, 0.1% (1 in 1,000) for blacks, and 0.5% (1 in 200) for Hispanics.

Researchers have suggested that the use of non-invasive methods in diagnosing melanoma requires extensive training unlike the use of naked eye. In other words, for a clinician to be able to analyze and interpret features and patterns derived from dermoscopic images, they must undergo through extensive training [7]. This explains why there is a wide gap between trained and untrained clinicians. Clinicians are often discouraged to use the naked eye as it has previously led to wrong diagnoses of melanoma. In fact, scholars encourage them to embrace routinely the use of portable automated real time systems since they are deemed to be very effective in prevention and early detection of melanoma [8].

### 1.1. Related Work

Skin image recognition on smart phones has become one of the attractive and demanding research areas in the past few years. Karargyris *et al.* have worked on an advanced image-processing mobile application for monitoring skin cancer [9]. The authors presented an application for skin prevention using a mobile device. An inexpensive accessory was used for improving the quality of the images. Additionally, an advanced software framework for image processing backs the system to analyze the input images. Their image database was small, and consisted of only 6 images of normal cases and 6 images of suspicious case.

Doukas *et al.* developed a system consisting of a mobile application that could obtain and recognize moles in skin images and categorize them according to their brutality into melanoma, nevus, and benign lesions. As indicated by the conducted tests, Support Vector Machine (SVM) resulted in only 77.06% classification accuracy [10].

Massone *et al.* introduced mobile teledermoscopy: melanoma diagnosis by one click. The system provided a service designed toward management of patients with growing skin disease or for follow-up with patients requiring systemic treatment. Teledermoscopy enabled transmission of dermoscopic images through e-mail or particular web-application. This system lacked an automated image processing module and was totally dependable on the availability of dermatologist to diagnose and classify the dermoscopic images. Hence, it is not considered a real-time system [11].

Wadhawan *et al*. proposed a portable library for melanoma detection on handheld devices based on the well-known bag-of-features framework [12]. They showed that the most computational intensive and time consuming algorithms of the library, namely image segmentation and image classification, can achieve accuracy and speed of execution comparable to a desktop computer. These findings demonstrated that it is possible to run sophisticated biomedical imaging applications on smart phones and other handheld devices, which have the advantage of portability and low cost, and therefore, can make a significant impact on health care delivery as assistive devices in underserved and remote areas. However, their system didn't allow the user to capture images using the smart phone.

Ramlakhan *et al.* [13] introduced a mobile automated skin lesion classification system. Their system consisted of three major components: image segmentation, feature calculation, and classification. Experimental results showed that the system was not highly efficient, achieving an average accuracy of 66.7%, with average malignant class recall/sensitivity of 60.7% and specificity of 80.5%. Barata *et al.* proposed two systems for the detection of melanoma cases in dermoscopy images using texture and color features [14]. The paper aimed at determining the best system for skin lesion classification. It was concluded that color features outperform texture features when used alone and that both methods achieve very good results, i.e., sensitivity = 96% and Specificity = 80% for global methods (i.e. global features color and texture) against Sensitivity = 100% and Specificity = 75% for local methods (i.e. local features color and texture). However, this system didn't run on a smart phone and didn't allow the users to capture skin images.

Sadeghi *et al.* [15] proposed an algorithm to identify the absence or presence of streaks in skin lesions, by analysing the appearance of detected streak lines. Using proposed features of the valid streaks along with the color and texture features of the entire lesion, an accuracy of only 76.1% was achieved.

Upon a careful review of literature, it is clearly observed that regular users, patients, and dermatologist can benefit from a portable application for skin cancer prevention and early detection.

## 2. SKINCURE APPLICATION OVERVIEW

SKINcure application is a smart phone-based application for iPhone or iPod with iOS 7.0 and onwards that will give the user live access to the current UV index and allow the user to calculate the time to skin burn with given parameters. The aspirant feature of this application is Dermoscopy Image Analysis that analyzes the dermoscopy skin images of the users and provides instantaneous results (i.e. classifies the image into normal, melanoma or atypical) using a remote image processing server.

The core functionalities of the SKINcure application are as follows:

1. Provide and show graphical representation of local UV status.
2. Calculate the time to skin burn and set notification alert.
3. Create and manage users mole images profile for dermoscopy analysis
4. Perform dermoscopy image analysis using a remote image processing server to classify the mole image into normal, atypical or melanoma.

The application is designed in a well-defined structure ensuring quality user experience for using the application features. In the following sections, the features of the new application is explored according to the screen/feature schematic design.

### 2.1. Current UV Screen

After starting up the application, the first screen where user will land is the Current UV screen. This screen has 3 modules as shown in Figure 1. First, the Location and temperature module with weather indicator presents information on current location and weather that gets updated with the weather condition. Second, the UV index module shows the UV index value of the current location. The index value is refreshed every 10 seconds. Third, the UV Status module provides the graph view of the UV index with color scale presentation mode. The horizontal axis is the time scale from 6 AM to 6 PM and the vertical axis is the UV index starting from 0 to 10+. This gives the user the standard UV index presentation to get a clear idea of UV index behaviour.

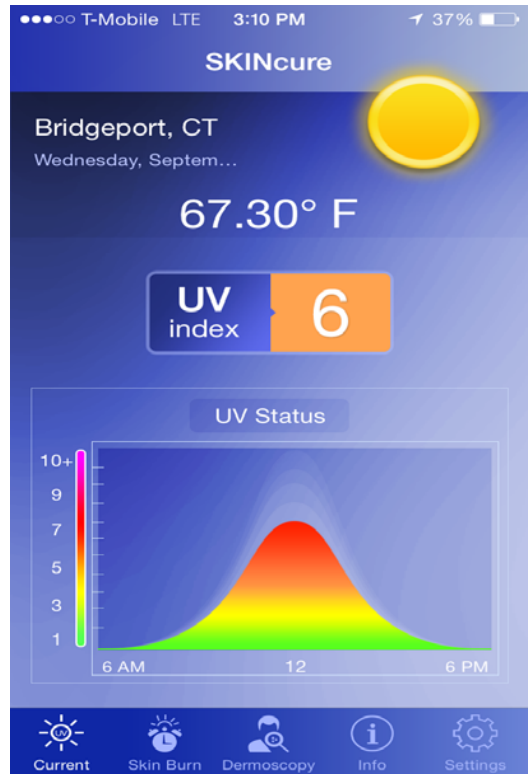

Figure 1. Current UV Screen

### 2.2. Time to Skin Burn Screen

The second tab screen is the "Time to Burn" as shown in Figure 2. This screen calculates the time to skin burn for given input set.

First, the set UV index input scroll bar that is auto set with current UV index allows the user to adjust as needed.

Second, the user can slide and select the environment type from the environment gallery box. The user can choose any option from building, shade, cloud, sand, wet sand, grass, wet grass, water and snow environments.

Third, from the skin type gallery view, the user selects skin type or the user can also tap on any skin type to enter the "Set Skin Type" screen to select any skin type. The usability of "Set Skin Type" will be explained in the next subsection.

Fourth, the user can set the SPF value ranging from 0 to 55+ using the horizontal scroll bar.

Finally, the "Estimated Time to Burn" is calculated for the selected properties. The Set Alarm button can be used to set notification alarm to let the user get the local notification from the application as the time is over.

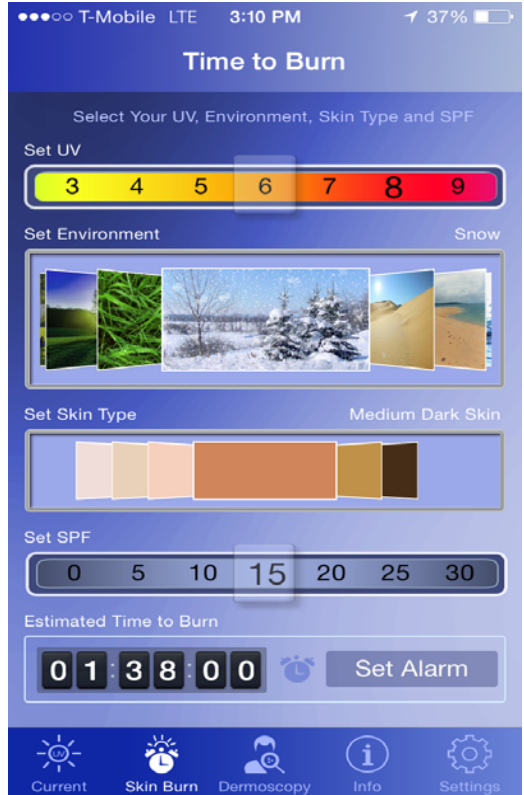
Figure 2: Time to Burn Screen

To calculate the time to skin burn in this application a model is created by deriving an equation to calculate the time for skin to burn namely, "*Time to Skin Burn*" (*TTSB*). This model is derived based on the information of burn frequency level and UV index level [16].

$$TTSB = TS / \begin{bmatrix} UV + (UV \times 0.85 \times SN) + \\ (UV \times 0.00487804 \times AL) + \\ (UV \times 0.2 \times SA) + \\ (UV \times 0.4 \times WSA) + \\ (UV \times 0.2 \times GR) + \\ (UV \times 0.4 \times WGR) + \\ (UV \times 0.15 \times BU) + \\ (UV \times 0.5 \times WA) - \\ (UV \times 0.5 \times SH) - \\ (UV \times 0.2 \times CL) \end{bmatrix} \times SPFW \quad (1)$$

*TS* is the time-to-skin-burn based on skin type where UV index equals to 1. Table 1 shows time-to-skin-burn at UV index of 1 for all skin types [17]. In Equation 1, *UV* is the ultraviolent index level ranging from 1 to 10, *AL* is the altitude in feet, *SN* represents snowy environment (Boolean value 0 or 1), *CL* represents cloudy weather (Boolean), *SA* represents sandy environment (Boolean), *WSA* represents wet sand environment (Boolean), *GR* represents grass environment (Boolean), *WGR* represents wet grass environment (Boolean), *BU* represents building environment (Boolean), *WA* represent shady environment (Boolean), *SH* represents water environment (Boolean) and *SPFW* is the sun protection factor weight. Table 2 shows

SPFW for various sun protection factor (SPF) levels [18]. The Boolean values are chosen based on the existence (1) or non-existence (0) of a certain weather condition or environment. The environmental factors indicate the amount of UV that a particular weather condition or environment reflects.

According to the TTSB model, a user with a skin type of 3 in a UV index of 10 at the sea level will start to experience sunburn just after 20 minutes of unprotected exposure to the sun (*TTSB=20 by substituting these values in Equation 1, UV =10, TS =200, SN =0, AL=0, SA=0, WSA=0, GR=0, WGR=0, BU=0, WA=0, SH=0, CL=0, SPFW=1*). If the user is using SPF15 sunscreen (*SPFW=3.7*), then the skin will start to burn after 74 minutes (*TTSB=74*). Continuing with the same example, if the user travels from the sea level to an altitude of 300 feet (*AL=300*) then the user's skin will start to experience sunburn after 30 minutes approximately (*TTSB=30*).

The proposed model can be validated by cross checking the calculated *TTSB values* from the previous example with the information provided by the National Weather Service Forecast [16]. Figure 3 shows the data provided by the National Weather Service and the calculated *TTSB values from our model are marked in red dots.* The calculated *TTSB* values fall in the range of the values provided by the National Weather Service. To the best of our knowledge, this is the first model proposed that calculates the time-to-skin-burn based on the given UV index, skin type, environmental parameters and SPF.

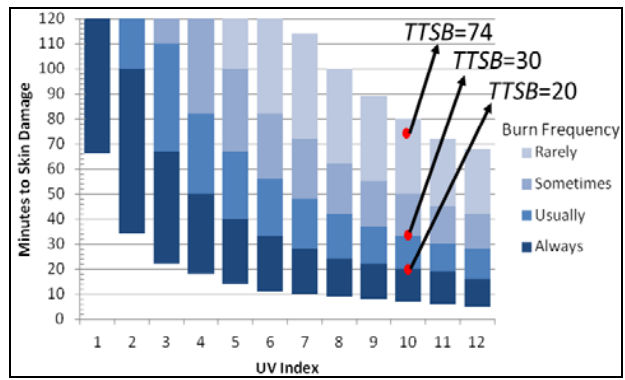

Figure 3. Flowchart for data provided by the National Weather Service and the calculated TTSB values

The environmental factor *AL*, will be automatically inserted into the model by detecting the user location using the smart phone GPS. Other factors such as UV, skin type, SPF levels, etc. will be manually selected by the user.

Table 1. Time to skin burn at UV index of 1 for all skin types

| Skin Type | Time to skin Burn at UV Index=1 (minutes) |
|---|---|
| 1 | 67 |
| 2 | 100 |
| 3 | 200 |
| 4 | 300 |
| 5 | 400 |
| 6 | 500 |

Table 2. SPFW for various (SPF) levels.

| SPF Level | SPFW |
|---|---|
| 0 | 1 |
| 5 | 1.3 |
| 10 | 2.4 |
| 15 | 3.7 |
| 20 | 4.5 |
| 25 | 4.8 |
| 30 | 7.5 |
| 35 | 8.2 |
| 40 | 9.5 |
| 45 | 11.3 |
| 50 | 12.4 |
| 50+ | 13.7 |

**2.2.1 Set Skin Type**

To browse through the skin types and to know the other properties like sensitivity and tendency, etc., the user can choose to be in this screen. In "Set Skin Type" screen; Figure 4; the user can check and select from the image gallery view for six skin type options such as Fair Light Skin, Light Skin, Medium Light Skin, Medium Dark Skin, Dark Skin and Deep Dark Skin. The image gallery has got sample skin types that matches with celebrities for ease of selection. The user can also browse the choices using the Skin Color Slider. This screen also presents the skin description for color and heritage as well as UV sensitivity and tendency to burn information for any selected option.

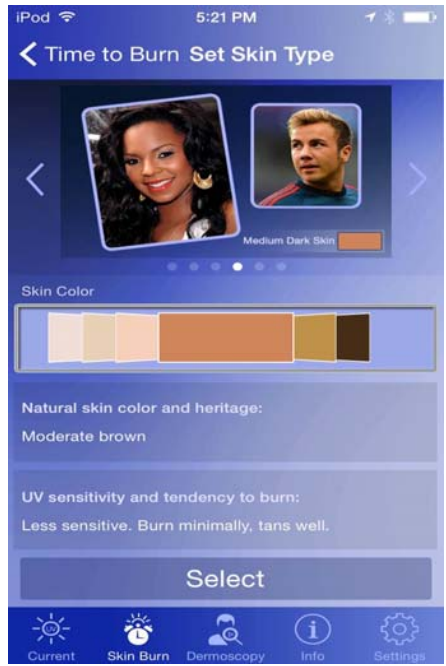

Figure 4. Set Skin Type Screen

## 2.3. Dermoscopy Screen

This screen gives the user the ability to manage different profiles. The user can create a new profile by using "+" button on top right corner. Each profile will be represented in a row with a little white circle that represents how many images are there in this profile. Initially, this image count is zero. The user may also select the "Edit" button on the top left corner of the app to delete any profile listed on the screen. Figure 5 shows different user profiles presented in a list. The user can select any profile to enter in and add new images in the selected profile. In following sub-section the "Profile" screen will be presented.

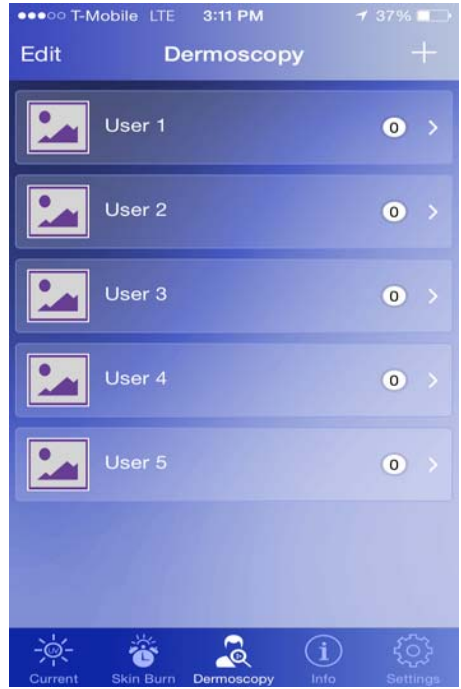

Figure 5: Profiles list in Dermoscopy Screen

### 2.3.1. Dermoscopy Profile Screen

Figure 6 (a) shows the Dermoscopy Profile Screen. This screen allows the user to capture dermoscopy images of the skin mole. The application will analyze the captured image and classify the mole into normal, atypical or melanoma. The application will ask the user to select the position of the mole on the skin as shown in Figure 6 (b) and (c).

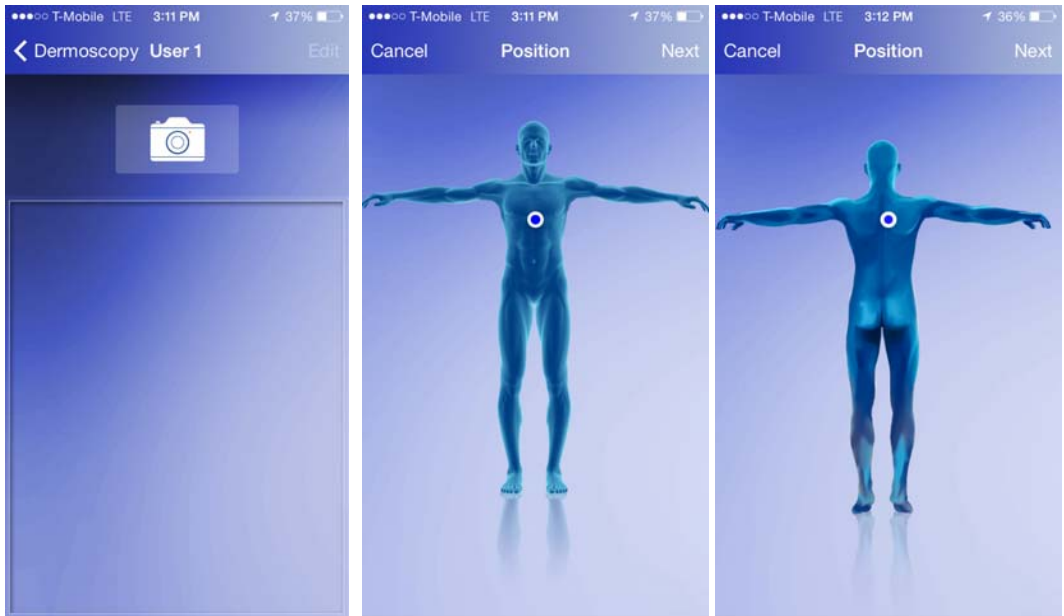

(a) (b) (c)

Figure 6. (a) Dermoscopy Profile Screen, (b) Dermoscopy Image Position Front Screen, (c) Dermoscopy Image Position Back Screen

**2.3.2. Dermoscopy Image View and Analysis Screen**

Figure 7 shows the "Dermoscopy Image View and Analysis" screen.

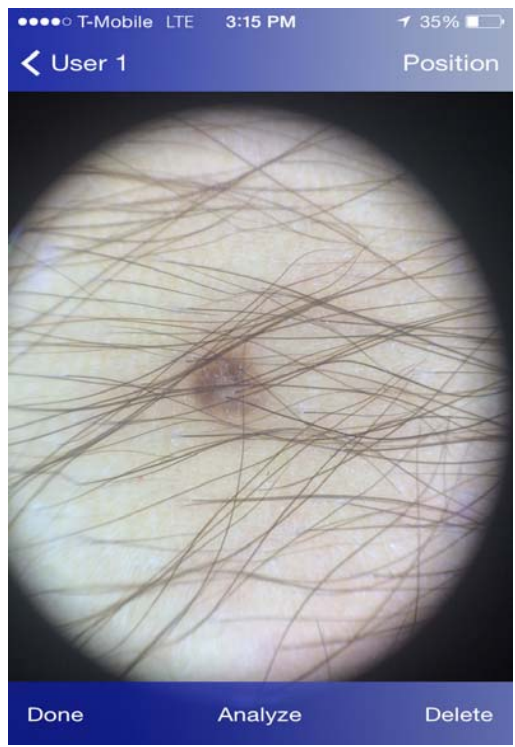

Figure 7. Dermoscopy Image Capture and Analyze Screen

The user can zoom in and out to have a clear look of the image in this screen. In addition, the user can also check the image position using "Position" button at top right of the screen. Finally, the user can tap on Analyze button to send the image to remote server to analyze the lesion and classify into normal, atypical or melanoma.

The image processing and classifications are done at the server side. The sever is located at the University of Bridgeport (UB), D-BEST lab, and thus the proposed system App does not require much processing power on the portable device side; only internet connection is needed to send the image to the server and receive the classification results. The system is important in the sense that it allows the users to detect melanoma at early stages which in turn increases the chance of cure significantly. Figure 8 shows the flow chart of the proposed dermoscopy image analysis system.

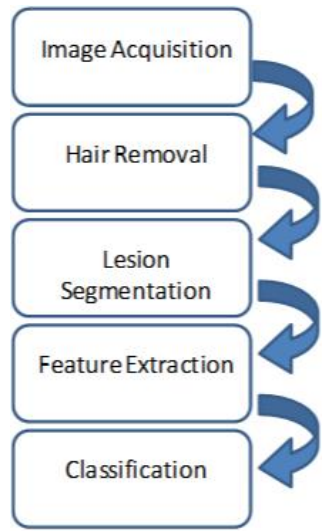

Figure. 8. Flowchart for the proposed dermoscopy image analysis system.

Using the iPhone camera solitary has some disadvantages since first, the size of the captured lesions will vary based on the distance between the camera and the skin, second, capturing the images in different light environments will be another challenge, and third, the details of the lesion will not be clearly visible. To overcome these challenges, a dermatoscope is attached to the iPhone camera. Figure 3 shows the dermatoscope device attached to the iPhone. The dermatoscope provides the highest quality views of skin lesions. It has a precision engineered optical system with several lenses. This provides the right standardized zoom with auto-focus and optical magnification of up to 20× directly to the camera of the iPhone device. Its shape ensures sharp imaging with a fixed distance to the skin and consistent picture quality. Also, it has a unique twin light system with six polarized and six white LEDs. This dermatoscope combines the advantages of cross-polarized and immersion fluid dermoscopy. Figure 9 shows samples of images captured using the dermatoscope attached to iPhone camera.

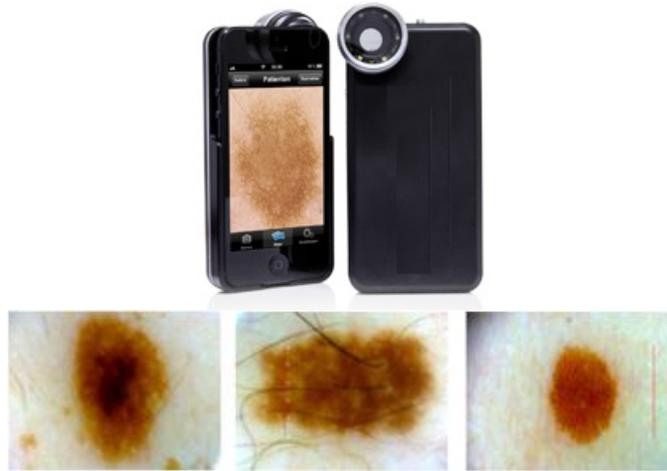

Figure 9. The dermatoscope device attached to the iPhone and sample of images captured using the device.

In dermoscopy images, if hair exists on the skin, it will appear clearly in the dermoscopy images. Consequently, lesions can be partially covered by body hair. Thus, hair can obstruct reliable lesion detection and feature extraction, resulting in unsatisfactory classification results. As a result, the image processing module in this application detect and exclude the hair from the dermoscopy images. The result is a clean image without hair. After that the lesion is segmented to separate the lesion from the background. This is an essential process before starting with the feature extraction.

In this work, five different feature sets are extracted. Feature extraction is the process of calculating parameters that represent the characteristics of the input image, whose output will have a direct and strong influence on the performance of the classification systems. These are 2-D Fast Fourier Transform, 2-D Discrete Cosine Transform, Complexity Feature Set, Color Feature Set and Pigment Network Feature Set. In addition to the five feature sets, the following four features are also calculated: Lesion Shape Feature, Lesion Orientation Feature, Lesion Margin Feature and Lesion Intensity Pattern Feature.

In the proposed system, the PH2 dermoscopic image database from Pedro Hispano hospital is used for the system development and for testing purposes [19]. The dermoscopic images were obtained under the same conditions using a magnification of 20×. This image database contains of a total of 200 dermoscopic images of lesions, including 80 normal moles, 80 atypical and 40 melanomas. They are 8-bit RGB color images with a resolution of 768×560 pixels. The images in this database is comparable to the images captured by the proposed system. We decided to use this database for testing purposes since it is verified and established by a group of dermatologists.

The extracted features are fed into a two level classifier, the first level classifier (classifier I) classifis the images into normal and abnormal, the senoncd level classifier (classifier II) classifis the abnormal images into atypical and melanoma.

Table 3 shows the confusion matrix for the two level classifier (classifier I and classifier II).

Table 3. Confusion matrix for the two level classifier
(classifier I and classifier II).

|  | Classifier I (%) | | Classifier II (%) | | |
|---|---|---|---|---|---|
|  | Normal | Abnormal | | | |
| Normal | 96.3 | 3.7 | | | |
| Abnormal | 2.5 | 97.5 | | Atypical | Melanoma |
|  |  |  | Atypical | 95.7 | 4.3 |
|  |  |  | Melanoma | 2.5 | 97.5 |

**2.4. Info Screen**

This screen provides general information about the UV index and Skin Cancer as well as operating techniques or procedures for the user of this application. The user will find a list of items in this screen and the user can tap any row and find the information screen on the selected topic. Figure 10 shows the info screen.

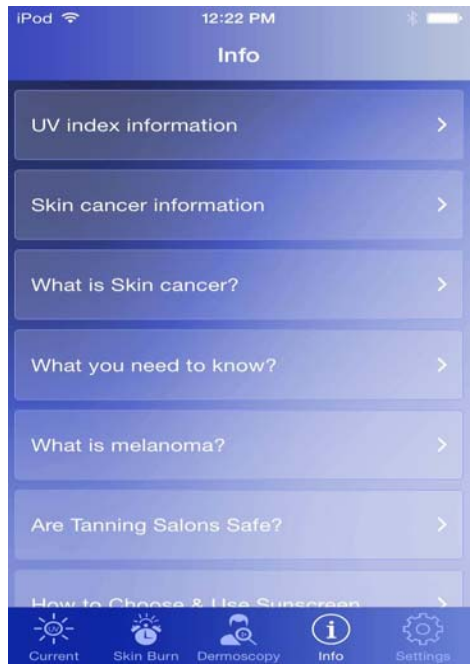

Figure 10: Information Screen

**2.5. Settings Screen**

The "Settings" screen, Figure 11, allows the user to set the general settings of the application such as Default Skin Type and UV Notification Alert.

To set up the Default Skin Type, the user slides through the skin type gallery view and selects skin type or the user can also tap on any skin type to enter the "Set Skin Type" screen to select any skin type. The features of "Set Skin Type" is explained in subsection 2.2.1. The selected skin type will be the default skin type for this application.

The UV notification alert option is another settings that allows the user to get notification as current UV index is found to be same or above UV index set threshold that the user can set in this screen. The user can set any value from 0 to 10 in "Set UV Index Threshold". The application will notify the user at the first time of the day whenever the current local weather reaches the UV index threshold value.

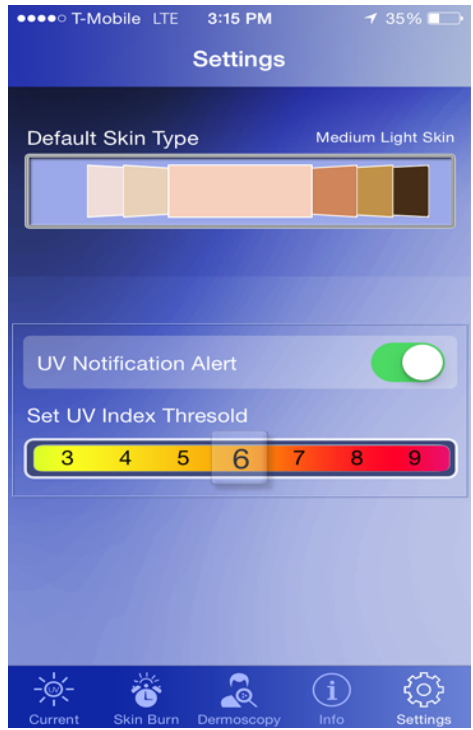

Figure 11. Settings Screen

## 3. CONCLUSIONS

This paper presented a fully functional smart phone-based application called SKINcure to aid in the malignant melanoma prevention and early detection. The proposed application has two major components. The first component is a real-time alert to help the users to prevent skin burn caused by sunlight. In this part, a novel equation to compute the time-to-skin-burn is introduced. The second component is an automated image analysis module where the user will be able to capture the images of skin moles and this image processing module classifies under which category the moles fall into; normal, atypical, or melanoma. An alert will be provided to the user to seek medical help if the mole belongs to the atypical or melanoma category. The proposed automated image analysis process includes image acquisition, hair detection and exclusion, lesion segmentation, feature extraction, and classification.

**Authors**

Omar Abuzaghleh (S'12) is a Ph.D. candidate majoring in Computer Science and Engineering at the University of Bridgeport, CT. He received the B.Sc. in Computer Science and Applications from the Hashemite University, Jordan in 2004, and M.S. degree in Computer Science from the School of Engineering, University of Bridgeport in 2007. He is a member of several professional organizations and specialty labs including; ACM, IEEE, the International Honor Society for The Computing and Information Disciplines (UPE), and the Digital/Biomedical Embedded Systems & Technology (D-BEST) Lab. Omar's current research interests include image processing, virtual reality, human computer interaction, cloud computing, and distributed database management.

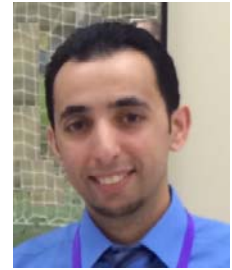

Miad Faezipour (S'06–M'10) is an Assistant Professor in the Computer Science & Engineering and Biomedical Engineering departments at the University of Bridgeport, CT and the director of the Digital/Biomedical Embedded Systems & Technology (D-BEST) Lab since July 2011. Prior to joining UB, she has been a Post-Doctoral Research Associate at the University of Texas at Dallas collaborating with the Center for Integrated Circuits and Systems and the Quality of Life Technology laboratories. She received the B.Sc. in Electrical Engineering from the University of Tehran, Tehran, Iran and the M.Sc. and Ph.D. in Electrical Engineering from the University of Texas at Dallas. Her research interests lie in the broad area of biomedical signal processing and behavior analysis techniques, high-speed packet processing architectures, and digital/embedded systems. Dr. Faezipour is a member of IEEE, EMBS and IEEE Women in Engineering.

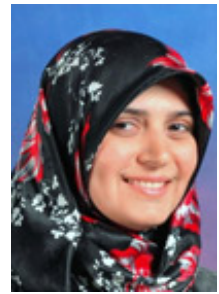

Buket D. Barkana (M'09) is an Associate Professor in the Electrical Engineering department at the University of Bridgeport, CT. Professor Barkana is also the director of the Signal Processing Research Group (SPRG) Laboratory in the Electrical Engineering program, University of Bridgeport. Areas of her research span all aspects of speech, audio, bio- signal processing, image processing and coding. She received the B.S. in Electrical Engineering from Anadolu University, Turkey, and the M.Sc. and Ph.D. in Electrical Engineering from the Eskisehir Osmangazi University, Turkey. Dr. Barkana is a member of IEEE.

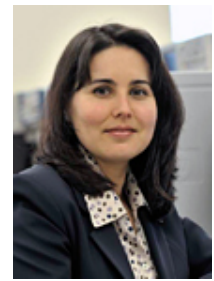